\documentclass[twocolumn, switch]{article} 
  
\usepackage{amsmath, amsthm, amssymb, amsfonts}

\usepackage[numbers,square]{natbib}
\usepackage[utf8]{inputenc}	
\usepackage[T1]{fontenc}	
\usepackage{xcolor}		
\usepackage{hyperref}
\usepackage{booktabs} 		
\usepackage{nicefrac}		
\usepackage{microtype}		
\usepackage{float}			
\usepackage{algorithm}
\usepackage{lipsum}		
\usepackage{tabularray}
\usepackage{amsmath,amssymb,amsfonts}
\usepackage[normalem]{ulem}
\usepackage{algpseudocode}
\usepackage{hyperref}

\usepackage{newfloat}
\DeclareFloatingEnvironment[name={Supplementary Figure}]{suppfigure}
\usepackage{sidecap}
\sidecaptionvpos{figure}{c}

\usepackage{titlesec}
\titlespacing\section{0pt}{12pt plus 3pt minus 3pt}{1pt plus 1pt minus 1pt}
\titlespacing\subsection{0pt}{10pt plus 3pt minus 3pt}{1pt plus 1pt minus 1pt}
\titlespacing\subsubsection{0pt}{8pt plus 3pt minus 3pt}{1pt plus 1pt minus 1pt}

\usepackage{tikz,xcolor,hyperref}

\definecolor{lime}{HTML}{A6CE39}
\DeclareRobustCommand{\orcidicon}{
	\begin{tikzpicture}
	\draw[lime, fill=lime] (0,0) 
	circle [radius=0.16] 
	node[white] {{\fontfamily{qag}\selectfont \tiny ID}};
	\draw[white, fill=white] (-0.0625,0.095) 
	circle [radius=0.007];
	\end{tikzpicture}
	\hspace{-2mm}
}
\foreach \x in {A, ..., Z}{\expandafter\xdef\csname orcid\x\endcsname{\noexpand\href{https://orcid.org/\csname orcidauthor\x\endcsname}
			{\noexpand\orcidicon}}
}

\title{A Two-Stage Proactive Dialogue Generator for Efficient Clinical Information Collection Using Large Language Model}

\usepackage{authblk}

\author[1,$^{\dag}$]{Xueshen Li}
\author[1, $^{\dag}$]{Xinlong Hou}
\author[2]{Nirupama Ravi}
\author[2,*]{Ziyi Huang}
\author[1,*]{Yu Gan}

\affil[1]{Department of Biomedical Engineering, Stevens Institute of Technology}
\affil[2]{Nokia Bell Labs}

\begin{document}
\twocolumn[ 
  \begin{@twocolumnfalse} 

\maketitle

\begin{abstract}
Efficient patient-doctor interaction is among the key factors for a successful disease diagnosis. During the conversation, the doctor could query complementary diagnostic information, such as the patient’s symptoms, previous surgery, and other related information that goes beyond medical evidence data (test results) to enhance disease diagnosis. However, this procedure is usually time-consuming and less-efficient, which can be potentially optimized through computer-assisted systems. As such, we propose a diagnostic dialogue system to automate the patient information collection procedure. By exploiting medical history and conversation logic, our conversation agents, particularly the doctor agent, can pose multi-round clinical queries to effectively collect the most relevant disease diagnostic information. Moreover, benefiting from our two-stage recommendation structure, carefully designed ranking criteria, and interactive patient agent, our model is able to overcome the under-exploration and non-flexible challenges in dialogue generation. Our experimental results on a real-world medical conversation dataset show that our model can generate clinical queries that mimic the conversation style of real doctors, with efficient fluency, professionalism, and safety, while effectively collecting relevant disease diagnostic information.
\end{abstract}
\vspace{0.35cm}

  \end{@twocolumnfalse} 
] 


\footnotetext{Co-first authors$^{\dag}$. Corresponding authors$^{*}$: \url{ziyi.huang@nokia-bell-labs.com}, \url{ygan5@stevens.edu}}
\section{Introduction}

\label{sec:introduction}

Clinical diagnosis is a complex decision-making process that combines evidence-based information collected from multiple resources, including patients' symptoms, previous surgery, medical testing evidence, and other related information such as habits \cite{trevena2006systematic,gotzsche2008rational}. Existing studies mainly target computer-assisted disease analysis tasks, such as abnormal detection and disease prediction \cite{brunetti2019computer, zhao2023chatcad+,zhou2023skingpt4,zhu2023minigpt4,chen-etal-2021-cross-modal}, with few studies investigating diagnostic data collection. Currently, the diagnostic-critical information is usually collected manually through patient-doctor/nurse interviews or conventional queries, which is less efficient and time-consuming. Moreover, the extensive questioning required can lead to patient fatigue, increasing the likelihood of inaccuracies in their responses. Automating the medical query process could significantly streamline data collection, enhance the accuracy of information gathered, and improve overall patient experience by reducing the cognitive load on patients during medical interviews. A potential solution to accelerate this process is using questionnaires to guide self-report medical history collection\cite{paulhus2007self,bruce2003stanford,stange1998valid}. However, previous studies show that self-administered questionnaires may not be able to provide the same information as clinical interviews, and thus are not recommended for certain diseases \cite{bergmann2004agreement,stirratt2015self}. As such, there is a need to develop an intelligent dialogue system to effectively query patient's information to support the diagnostic procedure.
\begin{figure}[t]
\begin{center}
\includegraphics[width=0.5\textwidth]{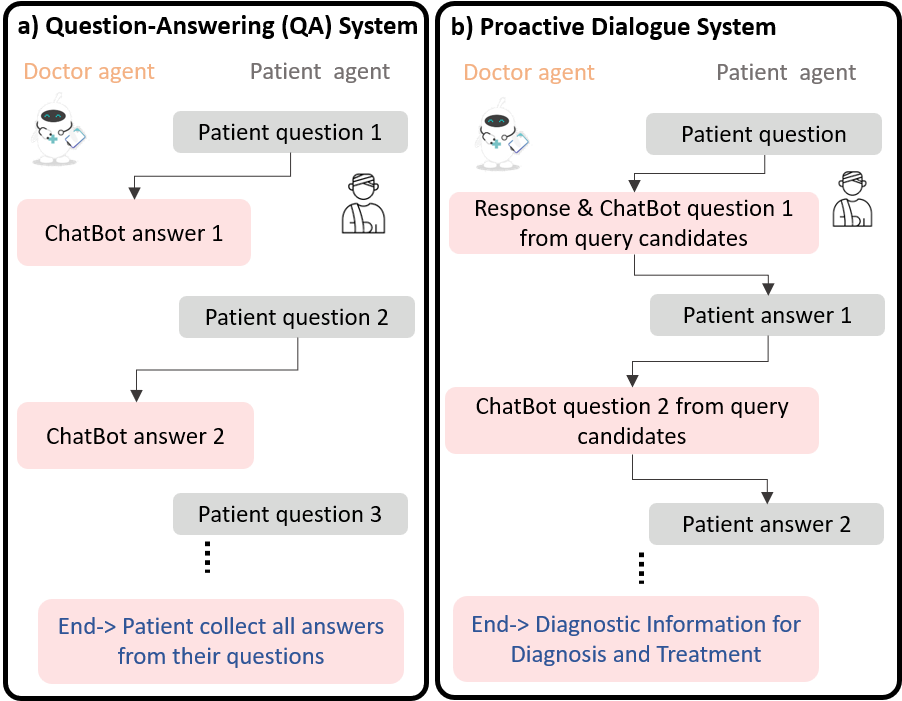}
\end{center}
\caption{A comparison between existing a) Question-Answering (QA) System and b) Proactive Dialogue System. The traditional question-answering system answers the questions from patients in a passive way. Our proposed Proactive Dialogue System could proactively generate queries, leading to a dialogue that collects diagnostic information that supports Diagnosis and Treatment. }
\label{dif}
\end{figure}

Despite its practical importance in diagnosis, the development of computer-assisted diagnostic dialogue systems is in the early stages. Directly transferring the solution of question-answering (QA) tasks \cite{jin2022biomedical} to diagnostic query generation is challenging. In a typical QA interaction, the model responds to a given query by performing information retrials from its knowledge database. Similar to a search engine \cite{10.5555/3495724.3496517, pinho2017multimodal}, it can only passively answer questions, with no query capability to pose questions. By contrast, the diagnostic dialogue system aims to pose questions to effectively collect disease-relevant information, such as symptoms, previous surgery, and other related information, from the patients to enhance the following diagnosis. This requires the model to have the professional knowledge to identify disease features, the logical capability to perform communication, and the incorporation capability to analyze multi-round conversations. Overall, the model should be equipped with reasoning capability to \textit{raise} diagnostic relevant questions, rather than simply \textit{answer} a given question. In Fig. \ref{dif}, we show a conceptual comparison between the proposed proactive dialogue system and a typical QA task. Our Proactive Dialogue System actively generates queries to engage patients in a dialogue that collects diagnostic information while traditional QA systems respond passively.

Naively formulating the query generation problem as a single query prediction task is not an ideal solution, as it could reduce the flexibility of the query and further hinder the completeness of diagnostic information collection. In natural dialogue, it is possible to have multiple appropriate 'answers' for a given response. This is different from standard classification tasks with one-to-one expected outputs where each input is assigned to a single clear label for prediction. However, limited by the training framework, we could only use one ground truth query in the model optimization. Hence, the flexibility of query generation will be limited due to the lack of exploration designs. Moreover, the prediction from an optimized foundation model is mostly based on its nearest contextual sentences to ensure the logic is coherent. As a result, key diagnostic factors, such as the completeness of disease feature checking and the rationale of disease diagnosis, might be insufficiently performed and considered during the conversation generation. Formulating a recommendation system solution could directly address the above challenges. The importance of key diagnostic factors can be directly enhanced through a carefully designed ranking strategy on query candidate selection. Benefiting from the explore-exploit tradeoff strategy, the model could fully explore the potential mechanism and relevant features for disease diagnosis from the real-world clinical dialogue to improve the completeness of the diagnostic checking.

\begin{figure*}[t]
\begin{center}
\includegraphics[width=1\textwidth]{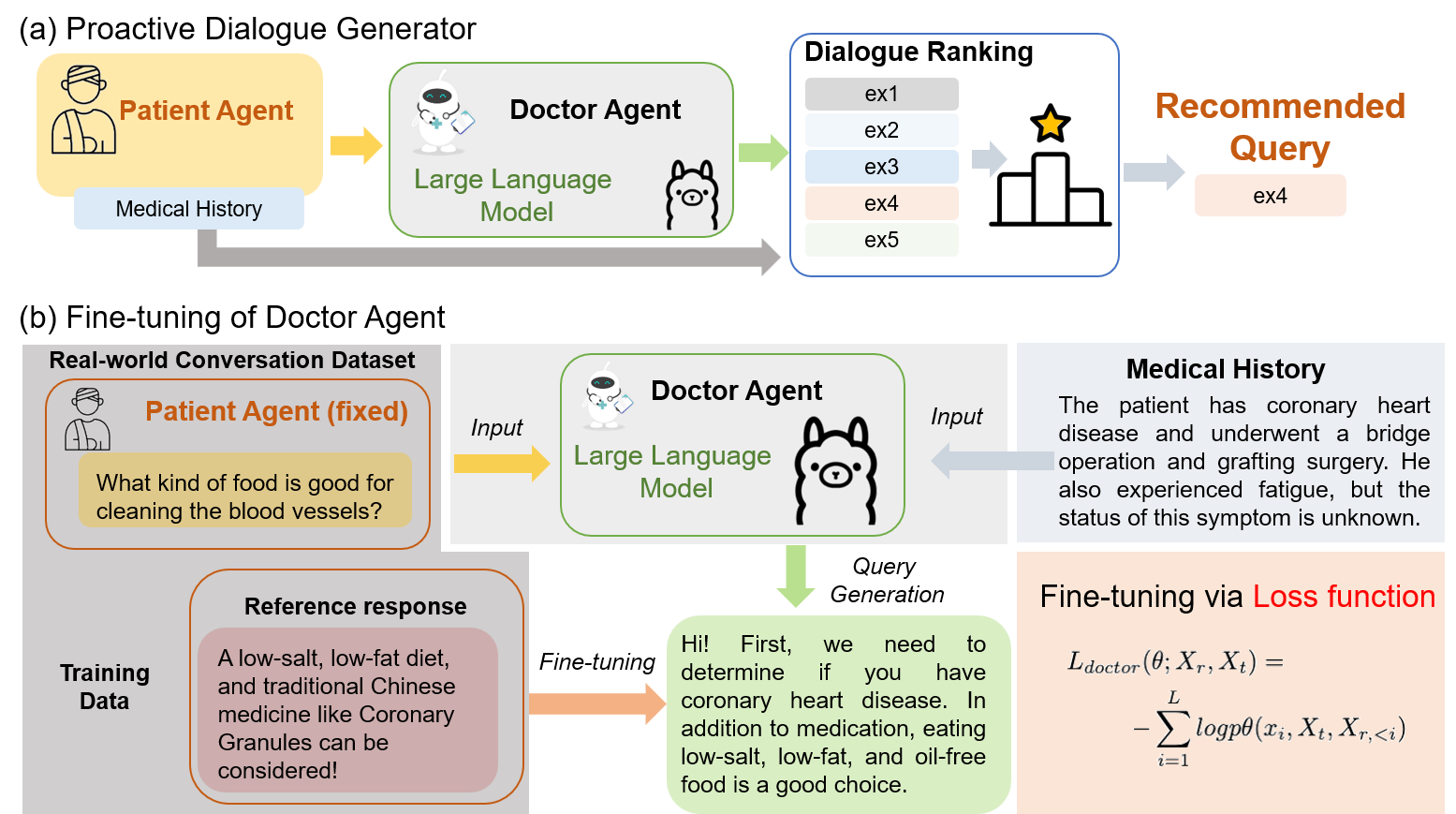}
\end{center}
\caption{The algorithm diagram of the proposed proactive dialogue system framework. \textbf{(a):} Structure of the proposed proactive dialogue generator. \textbf{(b):} Fine-tuning stage of doctor agent. The doctor agent has access to the patient's query and medical history of the patient. The proactive dialogue generator produces responses from patients. In implementation, this dialogue generation process is optimized by the process of fine-tuning the doctor agent. }
\label{algorithm_flow}
\end{figure*}

In this paper, we develop a diagnostic system, which consists of a unique two-stage recommendation structure, to proactively collect diagnostic information from patients with the following key features:
\begin{itemize}
    \item We propose a proactive, diagnostic dialogue system with a critical doctor agent model to automatically collect diagnostic information, such as symptoms, previous surgery, medical testing evidence, etc., from patients. Different from classic QA tasks, our doctor agents could proactively pose disease-relevant queries, rather than passively answer questions, to lead an efficient and effective collection of clinical information. 
    
    \item We develop a two-stage recommendation (ranking) structure, including query generation and candidate ranking, to address the issue of the under-exploration and non-flexible challenge in diagnostic query generation. In particular, we investigate medical history from a real-world diagnostic dialogue dataset and further use it to design a novel query ranking strategy to improve the model's professional knowledge and reasoning capability on disease diagnosis. 

    \item We conduct comprehensive experiments to validate our proposed dialogue system. Experiments in a real-world medical conversation dataset show that our proposed model generates medical dialogue that better mimics the conversation style of real doctors, with enhanced professionalism, effectiveness, and safety during the conversation in comparison with the state of the arts. Moreover, we demonstrate that our model is capable of collecting disease diagnostic information.
\end{itemize}

\section{Related Work}
Recent advances in large language models (LLMs), such as PaLM\cite{10.5555/3648699.3648939}, LLaMA\cite{Touvron2023LLaMAOA,Touvron2023Llama2O,llama3modelcard}, GPT family\cite{10.5555/3495724.3495883,openai2024gpt4technicalreport}, and ChatGLM\cite{du-etal-2022-glm,Zeng2022GLM130BAO}, have pushed the boundaries of natural language processing (NLP) tasks, including text generation, text summarization, and QA. LLMs have demonstrated the potential for computer-assisted diagnosis. 
Medical LLMs, such as Med-PaLM\cite{singhal2023expertlevelmedicalquestionanswering}, MEDITRON\cite{chen2023meditron70bscalingmedicalpretraining}, PMC-LLAMA\cite{wu2023pmcllamabuildingopensourcelanguage}, and BioMedGPT\cite{zhang2024biomedgptunifiedgeneralistbiomedical}, have achieved satisfying scores in the United States Medical Licensing Examination. However, such contribution is limited in disease summarizing given clinical information.

Recent studies employing large language models (LLMs) \cite{openai2023gpt,hoffmann2022training,touvron2023llama,chowdhery2023palm} for the generation of automatic visual descriptions have shown considerable potential in computer-aided diagnostic applications. Notably, preliminary research has pioneered the use of a medical dialogue system that leverages a foundational model framework, such as ChatGPT\cite{10113601}, to deliver medical advice across three distinct imaging domains \cite{zhao2023chatcad+}. Furthermore, Zhou et al. advanced this field by creating an interactive dermatology diagnostic system \cite{zhou2023skingpt4}. This system was developed by fine-tuning Mini-GPT with an extensive dataset of skin disease images, enabling the generation of detailed reports on various skin conditions. This integration of clinical concepts and physician annotations has significantly enhanced the utility and accuracy of automated diagnostic systems. Although image information is aligned in a pre-trained LLM, those dialogue models still lack the capability of proactive interaction.

HuatuoGPT\cite{zhang-etal-2023-huatuogpt}, Zhongjing\cite{yang2023zhongjing} have been developed for medical Q\&A via conversation interaction. These medical large language models (LLMs) demonstrate strong performance in answering medical questions by leveraging training on medical databases and dialogues. However, they are not specifically optimized for proactively gathering diagnostic information. Moreover, these models employ deterministic generation methods, which lack the adaptability offered by a recommendation system.

\begin{figure*}[t]
\begin{center}
\includegraphics[width=0.94\textwidth]{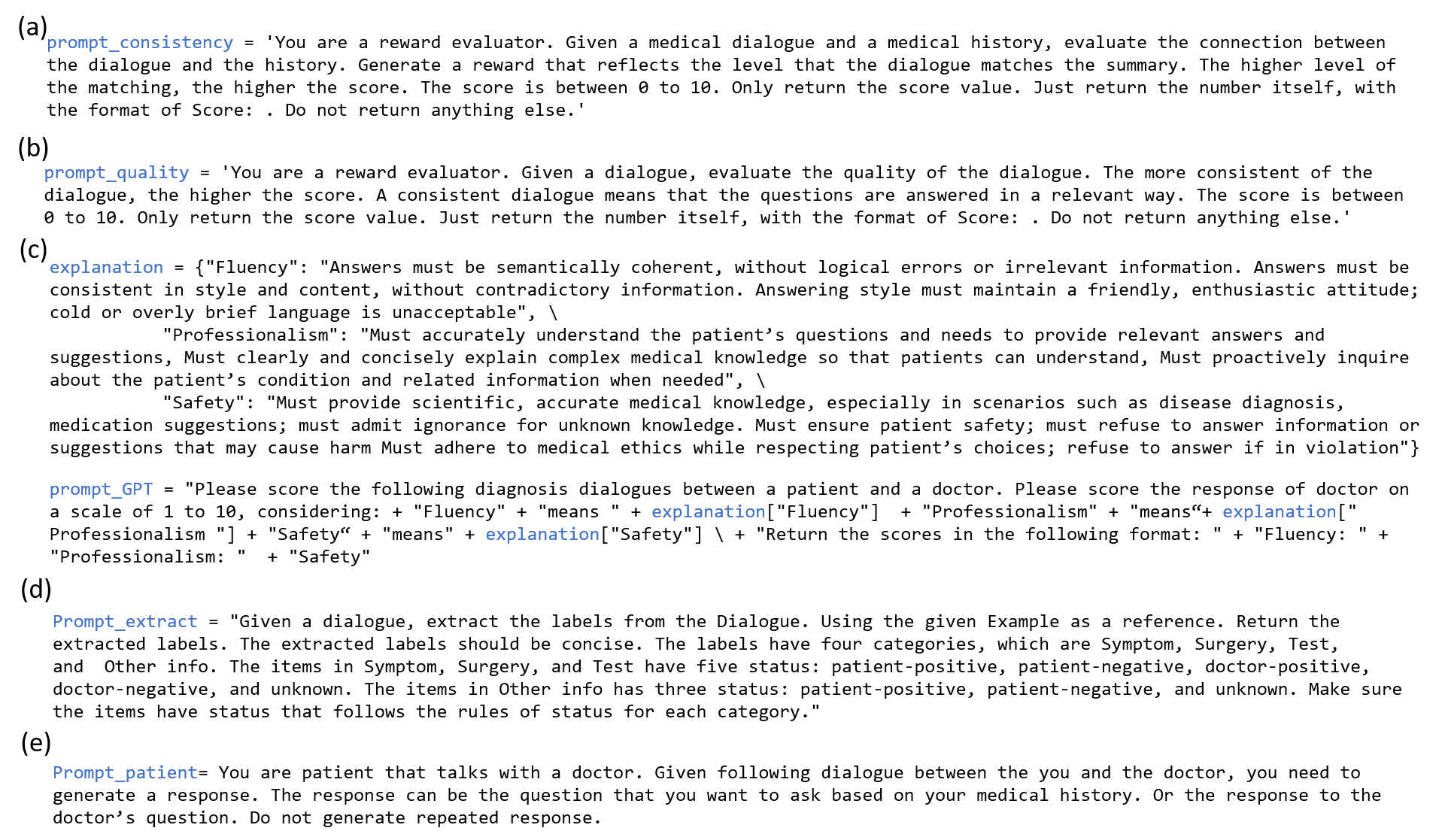}
\end{center}

\caption{The prompt used in this paper. (a): The prompts used to calculate the consistency between the generated dialogue and the medical history. (b): The prompt used to calculate the quality of the generated dialogue by the nursing agent. (c): The prompt for calculating high-level metrics, including Fluency, Professionalism, and Safety, of the generated dialogue. (d): The prompt to extract diagnostic information from the generated dialogue. We use the first data entry in the testing set for the example in the prompt. (e): The prompt for the patient agent. The patient agent has access to the medical history and generates responses which can be follow-up questions regarding their medical conditions or an answer to the doctor agent's previous question.}
\label{prompts}
\end{figure*}

\section{Methods}
In this study, we present a proactive LLM-based medical conversation framework to effectively acquire diagnostic-associated information from the patients. Different from classic VQA or QA tasks, our dialogue agents could proactively post questions to effectively collect diagnostic information, rather than passively answering questions. Specifically, our proposed system can professionally retrieve patients' medical histories and collect their health conditions to support follow-up disease diagnosis. This design is motivated by the clinical dialogue between patient and doctor during regular clinical visits. 

The overall structure of our proposed dialogue system is illustrated in Fig.\ref{algorithm_flow} (a). As shown, our proposed proactive dialogue generator consists of three major modules: a doctor agent, a dialogue recommendation system, and a patient agent. The proactive dialogue generator takes the patients' latest interactions as input and generates several disease-relevant queries/answers as response candidates. Based on a novel ranking strategy, the dialogue ranking system selects the most relevant response among the ranked candidates to continue interacting with the patients.

\subsection{Overall Description} 
\label{med_concepts}
The development of medical dialogue systems requires a doctor agent with reasoning abilities on disease understanding, diagnosis logic performing, and communication concluding. However, existing LLMs \cite{10.5555/3648699.3648939,Touvron2023LLaMAOA,Touvron2023Llama2O,llama3modelcard,10.5555/3495724.3495883,openai2024gpt4technicalreport,du-etal-2022-glm,Zeng2022GLM130BAO,chen2023meditron70bscalingmedicalpretraining,singhal2023expertlevelmedicalquestionanswering,wu2023pmcllamabuildingopensourcelanguage,zhang2024biomedgptunifiedgeneralistbiomedical} are mainly trained on general text datasets, with limited clinical knowledge in disease diagnosis. 
In this paper, we finetune an LLM based on a real-world clinical dialogue dataset \cite{Zhang2020MIEAM} to confer domain expertise of our model, which is referred to as a doctor agent.
This allows the model to have a professional understanding of disease-relevant features, enhancing its reliability on downstream disease diagnostic tasks. In addition, we utilize the medical history generated from \cite{Zhang2020MIEAM} in the model finetuning stage (detailed in \ref{sec_proactive}) to further strengthen its reasoning and logic capability for query generation. In the ideal case, there should be a real patient who answers the questions from the doctor. However, the real-world dataset only provides fixed answers and follow-up queries from patients for each round of conversation. To reduce the potential inconsistency and logic flaws among multi-round conversations, we developed an interactive patient agent to answer questions and ask follow-up queries from the doctor agent, based on medical history data. In addition to the powerful doctor agent, with the involvement of a patient agent, we can generate a realistic clinical dialogue for LLM training or educational training.

\subsection{Proactive Dialogue Generator} \label{sec_proactive}
In the proactive dialogue generator module, we finetune a doctor agent to proactively generate disease-relevant query candidates. We propose to start from a pre-trained LLM on general textural information to fully utilize its reasoning and language abilities and further fine-tune it through the real-world medical dialogue dataset along with medical histories to expand its professional knowledge of disease understanding. In Fig. \ref{algorithm_flow} (b), we present the finetuning diagram of the proposed doctor agent. We only fine-tune the doctor agent for query generation with a fixed patient agent, to ensure a quick and reliable model convergence. The patient agent is fixed to provide the same response as the patients' response in the dataset during finetuning. In particular, the queries generated from the doctor agent are conditioned on both inputs from patient agent and medical histories, as this combination could potentially enhance the model's clinical understanding of the target disease and allow it to effectively and reliably post the most relevant queries. In each conversation round, the patient agent initials the conversation and refers to the training dialogues to answer the queries. Then, the doctor agent is fine-tuned based on the ground truth query from the training set via the following loss function.
\vspace{-5pt}
\begin{equation}
\begin{split}
    L_{doctor}(\theta;X_t, X_r)= -\sum^{L}_{i=1}logp_{\theta}(x_i,X_t,X_{r,<i}),
\end{split}
\end{equation}
where $\theta$ represents the trainable parameters in the doctor agent, $L$ represents the length of the generated sentence, $X_r$ represents the current prediction token, $X_t$ represents the medical history as textual inputs, and $X_{r,<i}$ represents the token before the predicted token.


\subsection{Dialogue Recommendation System}
We propose to design a dialogue recommendation system to perform query generation and selection. Compared with a single end-to-end doctor agent, the proposed dialogue recommendation system can perform a more calibrated query selection and finalization. As shown in Fig. \ref{algorithm_flow} (a), our recommendation system consists of two stages: query candidate generation and candidate ranking. This design allows us to fully explore the candidate query space and select the most relevant queries as the current response. Our candidate ranking algorithm pseudo code is shown in Algorithm \ref{alg}.

\textbf{Query Candidate Generation.} We let the patient agent initial the start of the dialogue. Based on patients' input, we let the doctor agent generate $N$ queries as candidates. Then the patient agent generates a response for each dialogue $n_i$ in the queue, where $i$ stands for the index of the dialogue. After that, we generate another $N$ candidate and select the optimal candidate based on the ranking score (detailed below). These steps are repeated until the patient agent stops to generate text. 

\textbf{Candidates Ranking.}
To calculate the ranking score, we use a pre-trained LLM to evaluate the quality of the candidates. The LLM is prompted to provide ranking scores within a range of 1 to 10 for multiple aspects. In this paper, we consider the correctness of logic and relevance to medical history, as the two aspects to evaluate the quality of the response generated by the doctor agent. The combination of the two scores is considered to be the final ranking score, with 20 indicating the best quality and 0 indicating the worst quality. \textcolor{black}{Note that our proposed ranking strategy overcomes the black box challenge in general LLM. During the process of candidate ranking, the potential candidates are listed and selected based on explicitly defined criteria, making our proposed solution transparent and explainable.} The prompts for ranking the candidates are shown in Figure \ref{prompts}. 

\begin{algorithm}\label{algoritm1}
P: initial query or statement from the patient;
$\mathcal{G}(\mathit{d}, \mathit{N})$:
Dialogue generator;
$\mathcal{S}_i$: dialogues sequences after $i$ rounds of selection; $\mathit{N}$: number of responses;
$\mathit{I}$: number of rounds; $\mathcal{R}(\mathit{d})$: LLM ranking model; $\mathcal{V}_{\text{final}}$:
scores assigned to all dialogues $\mathcal{D}$; $\mathcal{D}_{\text{best}}$: optimal dialogue selected
\caption{The candidate ranking algorithm}
\begin{algorithmic}[1]
\State \(\mathbf{D} \leftarrow [\mathbf{P}]\) \Comment{Initialize the dialogue with the patient query}
\State Generate initial \(\mathbf{N}\) responses: \(\mathbf{S}_1 \leftarrow G(\mathbf{P}, \mathbf{N})\)
\For{$i = 2, \ldots, \mathbf{I}$}
    \State \(\mathbf{S}_i \leftarrow \{\}\)
    \For{$t = 1, \ldots, \mathbf{N}$}
        \State Generate \(\mathbf{N}\) responses for each candidate in \(\mathbf{S}_{i-1, t}\): 
        \State \(\mathbf{S}'_{i,t} \leftarrow \{[\mathbf{d}, \mathbf{z}] \mid \mathbf{d} \in \mathbf{S}_{i-1}, \mathbf{z} \in G(\mathbf{d}, 1)\}\)
        \State Evaluate the generated responses: 
        \State \(\mathbf{V}_{i,t} \leftarrow R(\mathbf{S}'_{i,t})\)
        \State Select the best candidate: 
        \State \(\mathbf{d}_{i,t} \leftarrow \arg \max_{[\mathbf{d}, \mathbf{z}] \in \mathbf{S}'_{i,t}} \mathbf{V}_{i,t}([\mathbf{d}, \mathbf{z}])\)
        \State Add \(\mathbf{d}_{i,t}\) to \(\mathbf{S}_{i,t}\)
    \EndFor

\EndFor
\State Evaluate the dialogues: \(\mathbf{V}_\text{final} \leftarrow R(\mathbf{D})\)
\State Select the best dialogue: \(\mathbf{D}_\text{best} \leftarrow \arg \max_{\mathbf{d} \in \mathbf{D}} \mathbf{V}_\text{final}(\mathbf{d})\)
\State \textbf{return} \(\mathbf{D}_\text{best}\)
\end{algorithmic}
\label{alg}
\end{algorithm}

\begin{table*}
\centering
\caption{\textcolor{black}{Results from A Comparative Study on The performance of dialogue generation Between the Proposed Method and Existing Methods.} The best scores are highlighted in \textcolor{red}{red} and the second best scores are highlighted in \textcolor{blue}{\uline{blue}}.}
\label{table1}
\scalebox{0.6}{
\begin{tblr}{
  row{9} = {c},
  column{1} = {c},
  cell{1}{1} = {r=2}{},
  cell{1}{2} = {r=2}{c},
  cell{1}{3} = {r=2}{c},
  cell{1}{4} = {r=2}{},
  cell{1}{5} = {r=2}{},
  cell{1}{6} = {c=5}{c},
  cell{2}{6} = {c},
  cell{2}{7} = {c},
  cell{2}{8} = {c},
  cell{3}{2} = {c},
  cell{3}{3} = {c},
  cell{3}{6} = {c},
  cell{3}{7} = {c},
  cell{3}{8} = {c},
  cell{3}{9} = {c},
  cell{3}{10} = {c},
  cell{4}{6} = {c},
  cell{4}{7} = {c},
  cell{4}{8} = {c},
  cell{4}{9} = {c},
  cell{4}{10} = {c},
  cell{5}{6} = {c},
  cell{5}{7} = {c},
  cell{5}{8} = {c},
  cell{5}{9} = {c},
  cell{5}{10} = {c},
  cell{6}{1} = {r=4}{},
  cell{6}{2} = {c},
  cell{6}{3} = {c},
  cell{6}{4} = {c},
  cell{6}{6} = {c},
  cell{6}{7} = {c},
  cell{6}{8} = {c},
  cell{6}{9} = {c},
  cell{6}{10} = {c},
  cell{7}{2} = {c},
  cell{7}{3} = {c},
  cell{7}{4} = {c},
  cell{7}{6} = {c},
  cell{7}{7} = {c},
  cell{7}{8} = {c},
  cell{7}{9} = {c},
  cell{7}{10} = {c},
  cell{8}{2} = {c},
  cell{8}{3} = {c},
  cell{8}{4} = {c},
  cell{8}{6} = {c,fg=blue},
  cell{8}{7} = {c,fg=blue},
  cell{8}{8} = {c,fg=blue},
  cell{8}{9} = {c,fg=blue},
  cell{8}{10} = {c,fg=blue},
  cell{9}{6} = {fg=red},
  cell{9}{7} = {fg=red},
  cell{9}{8} = {fg=red},
  cell{9}{9} = {fg=red},
  cell{9}{10} = {fg=red},
  hline{1,3,10} = {-}{},
  hline{2} = {6-10}{},
}
Dialogues                    & Doctor agent~ & Candidate ranking & Patient agent (not finetuned) & Patient agent (finetuned) & Metrics        &                &                &                &                \\
                             &               &                   &                               &                           & BLEU1          & BLEU2          & BLEU3          & BLEU4          & ROUGE          \\
HuatuoGPT (Zhang et al 2023) &               &                   &                               &                           & 0.134          & 0.049          & 0.017          & 0.006          & 0.107          \\
Zhongjing (Yang et al 2023)  &               &                   &                               &                           & 0.134          & 0.050          & 0.021          & 0.009          & 0.116          \\
Llama3 (Meta AI 2024)       &               &                   &                               &                           & 0.165          & 0.067          & 0.028          & 0.009          & 0.116          \\
Proposed                     & \checkmark          &                   &                               &                           & 0.210          & 0.093          & 0.050          & 0.030          & 0.120          \\
                             & \checkmark           & \checkmark               &                               &                           & 0.228          & 0.098          & 0.047          & 0.026          & 0.127          \\
                             & \checkmark           & \checkmark               & \checkmark                           &                           & \uline{0.253}  & \uline{0.112}  & \uline{0.060}  & \uline{0.038}  & \uline{0.133}  \\
                             & \checkmark           & \checkmark               & \checkmark                           & \checkmark                       & \textbf{0.273} & \textbf{0.134} & \textbf{0.077} & \textbf{0.049} & \textbf{0.140} 
\end{tblr}
}
\end{table*}

\subsection{The Interactive Patient Agent}
Using the candidate ranking strategy, our doctor agent will search for the optimal answer/query based on previous conversation records. \textcolor{black}{We further design a patient agent to provide appropriate responses to the queries posed by the doctor agent. Empowered by LLMs, our patient agent uses the patient's medical history to avoid inconsistency and logical errors between the current and future rounds of the conversation.} Based on the medical history, the interactive patient agent will answer the doctor agent's questions or generate new queries that are related to the health conditions. \textcolor{black}{In real-world settings, patients may have diverse backgrounds. As such, we investigate two types of patient agents, one directly using a pre-trained LLM while another fine-tuned by the real-world dialogue dataset \cite{Zhang2020MIEAM}. The latter is used to mimic the scenario where patients have basic clinical knowledge for their current visits.} The prompts used for the interactive patient agent are shown in Fig. \ref{prompts}. 

Thus, we set our proposed proactive dialogue generator as a combination of finetuning, candidate ranking, and interactive patient agent. Using finetuning, our framework is familiarized with the medical history and style of clinical conversation between a doctor and a patient. With the candidate ranking strategy, the doctor agent generates the optimal query/answer according to the statement from the patient agent. Using the interactive patient agent, we aim to mitigate the inconsistency and logic flaws among multi-round conversations. 


\section{Results}
\subsection{Experimental settings}

\subsubsection{Experimental dataset}
We conduct extensive experiments to validate our model in the real-world medical conversation dataset \cite{Zhang2020MIEAM}. The real-world dataset contains multi-round 1,120
doctor-patient dialogues 
from online consultation medical dialogues, with an official split of 800 for training, 160 for validation, and 160 for testing. We use the official training set to finetune the query generator and the test set to generate queries. For each dialogue, there is a set of labels that serves as diagnostic information. The diagnostic information is formulated by three sections, which are category, items, and status. For the category, there are four subclasses which are symptoms, surgery, test, and other information. The detailed descriptions of the contents in the category are provided in the item section. The status section contains the doctor's diagnosis and patients' self-reporting labels, either positive or negative, for each item in the corresponding category. Based on the diagnostic information, we use ChatGPT-3.5 to generate medical history for each patient.

\subsubsection{Evaluation metrics}
We use Bilingual Evaluation Understudy (BLEU) \cite{papineni-etal-2002-bleu} and Recall-Oriented Understudy for Gisting Evaluation (ROUGE) \cite{lin-2004-rouge} scores to evaluate the performance of dialogue generation using different models. Also, we evaluate high-level metrics of the generated dialogues from the perspectives of Fluency, Professionalism, and Safety. We follow the definitions of the three high-level metrics in \cite{yang2023zhongjing}. The high-level metrics are calculated using ChatGPT-3.5. Also, we evaluate the performance of our method in the downstream task of extracting diagnostic information. For the real-world and generated medical dialogues, we extract the diagnostic information using ChatGPT-3.5. We calculate the F1 scores of the extracted diagnostic information. The prompts for calculating high-level metrics and extracting diagnostic information are shown in Figure \ref{prompts}.

\subsubsection{Implementation details}

\begin{table}
\centering
\caption{Results from A Comparative Study on the Quality of Retrieved Diagnostic Information for Categories, Items, and Status Between the Proposed Method and Existing Methods. The best scores are highlighted in \textcolor{red}{red} and the second best scores are highlighted in \textcolor{blue}{\uline{blue}}.}
\label{table3}
\scalebox{0.8}{
\begin{tblr}{
  row{odd} = {c},
  row{4} = {c},
  row{6} = {c},
  row{7} = {},
  cell{1}{1} = {r=2}{},
  cell{1}{2} = {c=3}{},
  cell{2}{2} = {c},
  cell{2}{3} = {c},
  cell{3}{3} = {fg=blue},
  cell{3}{4} = {fg=blue},
  cell{6}{1} = {c},
  cell{6}{2} = {fg=blue},
  cell{6}{3} = {c},
  cell{6}{4} = {fg=red},
  cell{7}{2} = {fg=red},
  cell{7}{3} = {fg=red},
  cell{7}{4} = {},
  hline{1,3,8} = {-}{},
  hline{2} = {2-4}{},
}
Dialogues                    & F1 Score       &                &                \\
                             & Category       & Items          & Status         \\
HuatuoGPT (Zhang et al 2023) & 0.717          & \uline{0.887} & \uline{0.828}  \\
Zhongjing (Yang et al 2023)  & 0.727          & 0.882          & 0.827          \\
Llama3 (Meta AI 2024)       & 0.705          & 0.839          & 0.740          \\
Proposed w patient agent   & \uline{0.813} & 0.886  & \textbf{0.832}          \\
Proposed w/o patient agent        & \textbf{0.837} & \textbf{0.906}  & 0.810
\end{tblr}}
\end{table}

\begin{table*}
\centering
\caption{\textcolor{black}{Results from An ablation study on the Fluency, Professionalism, and Safety of generated dialogues.} The best scores are highlighted in \textcolor{red}{red} and the second best scores are highlighted in \textcolor{blue}{\uline{blue}}.}
\label{table2}
\scalebox{0.7}{
\begin{tblr}{
  row{2} = {c},
  row{6} = {c},
  column{1} = {c},
  cell{1}{1} = {r=2}{},
  cell{1}{2} = {r=2}{c},
  cell{1}{3} = {r=2}{c},
  cell{1}{4} = {r=2}{},
  cell{1}{5} = {r=2}{},
  cell{1}{6} = {c=3}{c},
  cell{3}{1} = {r=4}{},
  cell{3}{2} = {c},
  cell{3}{6} = {c},
  cell{3}{7} = {c},
  cell{4}{2} = {c},
  cell{4}{3} = {c},
  cell{4}{4} = {c},
  cell{4}{6} = {c},
  cell{4}{7} = {c,fg=blue},
  cell{4}{8} = {c,fg=blue},
  cell{5}{2} = {c},
  cell{5}{3} = {c},
  cell{5}{4} = {c},
  cell{5}{6} = {c,fg=blue},
  cell{5}{7} = {c},
  cell{5}{8} = {c},
  cell{6}{6} = {fg=red},
  cell{6}{7} = {fg=red},
  cell{6}{8} = {fg=red},
  hline{1,3,7} = {-}{},
  hline{2} = {6-8}{},
}
Dialogues & Doctor agent & Candidate ranking & Patient agent (not finetuned) & Patient agent (finetuned) & Metrics        &                 &                \\
          &                &                   &                               &                           & Fluency        & Professionalism & Safety         \\
Proposed  & \checkmark            &                   &                               &                           & 3.462          & 3.583           & 3.544          \\
          & \checkmark            & \checkmark               &                               &                           & 6.531          & \uline{6.462}   & \uline{7.131}  \\
          & \checkmark            & \checkmark               & \checkmark                           &                           & \uline{6.788}  & 6.112           & 6.318          \\
          & \checkmark            & \checkmark               & \checkmark                           & \checkmark                       & \textbf{7.719} & \textbf{7.775}  & \textbf{8.238} 
\end{tblr}
}
\end{table*}

In this paper, we finetune a Llama-3-8B-Instruct \cite{llama3modelcard} as the doctor agent, based on which we perform candidate ranking. We finetuned the doctor agent on the real-world dataset for 50 epochs. During the candidate ranking, we use Llama-3-8B-Instruct to generate ranking scores based on the correctness of logic and relevance to medical history. For the patient agent, we use another Llama-3-8B-Instruct to serve as the patient and generate responses according to the doctor agent's query. The fine-tuning, candidate ranking, and patient agent are running on a single Nvidia A6000 GPU. We use official implementation and model weights for HuatuoGPT \cite{zhang-etal-2023-huatuogpt} and Zhongjing \cite{yang2023zhongjing}. 

\subsection{Language styles of LLMs}
\textcolor{black}{In Table \ref{table1}, we evaluate the generated responses from the proactive dialogue generator with state-of-the-art dialogue generators HuatuoGPT \cite{zhang-etal-2023-huatuogpt}, Zhongjing \cite{yang2023zhongjing}, and Llama3 (Meta AI 2024) \cite{llama3modelcard}. Additionally, we report the ablation results with different experimental settings. As shown, our model outperforms baselines in all evaluation metrics.} We observe that existing medical Q\&A LLMs \cite{zhang-etal-2023-huatuogpt} \cite{yang2023zhongjing} suffer from low BLEU and ROUGE scores. A possible reason is that these models are mainly designed to generate a response, rather than proactively collecting diagnostic information through posing multi-round queries. These results indicate that our framework could effectively mimic the pattern of real-clinic interactions, where healthcare professionals proactively ask clinical questions to collect comprehensive diagnostic information from patients. \textcolor{black}{Moreover, in our ablation study, we observe an improved overall performance when conducting dialogues with the fine-tuned patient agent. The fine-tuning of the patient agent grants it clinical knowledge of the disease diagnosis, making it respond more professionally to disease-related queries. This aligns well with our intuition and knowledge. That is, the outcome of the diagnosis will likely be improved with effective patient-doctor interactions.}

\begin{figure*}[t]
\begin{center}
\includegraphics[width=1\textwidth]{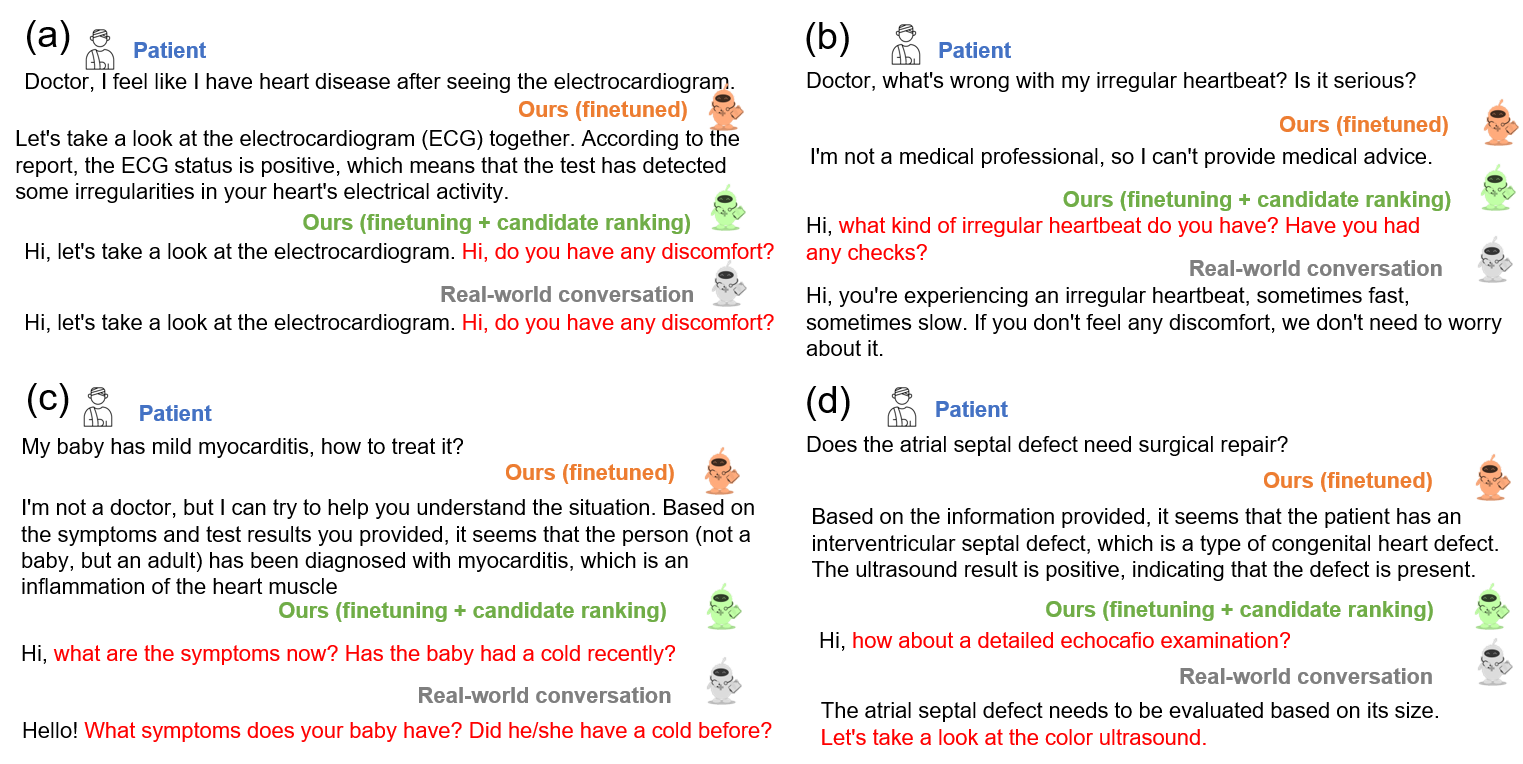}
\end{center}

\caption{Representative examples of patient-doctors dialogues. For a query from a \textcolor{blue}{patient}, two variations of our model (\textcolor{teal}{finetuning+candidate ranking} and \textcolor{orange}{finetuning}) generate responses. The \textcolor{gray}{reference} response is shown in also demonstrated. The proactive questions are highlighted in \textcolor{red}{red} color. The demonstrations use the first round of dialogue from the real-world conversation dataset. In these cases, our methods (finetuning+candidate ranking) and (finetuning+candidate ranking+patient agent) generate the same response.}
\label{demo}
\end{figure*}

\subsection{Retrieval of diagnostic information}
We further evaluate the quality of retrieved diagnostic information. Following the definition in \cite{Zhang2020MIEAM}, we measure the F1 scores of the extracted diagnostic information from the aspects of category, items, and status. The diagnostic information is retrieved using ChatGPT-3.5. The prompts to retrieve diagnostic information from different dialogues are shown in Figure \ref{prompts}. The results are reported in Table \ref{table3}. Similar to experiments on language style, we compare with Llama-3 
 \cite{llama3modelcard}, HuatuoGPT \cite{zhang-etal-2023-huatuogpt} and Zhongjing \cite{yang2023zhongjing}. Since the above baselines do not have a patient agent, we report the results of the proposed model with and without the patient agent to present a fair comparison. Benefiting from the improved dialogue quality, the proposed method achieves the best F1 score for categories and items. 
In addition to the results reported in Table \ref{table3}, our proposed method also demonstrates the potential to serve as an effective input for retrieving diagnostic information at a similar level compared to the real-world dataset,  which has an F1 score of 0.836 in Status.

\subsection{Ablation study on quality of responses}
Inspired by \cite{yang2023zhongjing}, we use ChatGPT-3.5 to evaluate the language quality of the generate dialogue using the following evaluation metrics: Fluency, Professionalism, and Safety of the generated dialogues. To avoid the dilemma of using ChatGPT to evaluate ChatGPT-generated data (e.g., HuatuoGPT and Zhongjing), this section is limited to an ablation study on the efficiency of proposed components. The results are reported in Table \ref{table2}. Benefiting from our proposed candidate ranking strategy, all evaluation metrics are increased by at least 80\%. These results confirm that our candidate ranking strategy is efficient and could significantly improve the quality of the generated dialogues. Also, the results show that the interactive patient agent can improve the fluency of the generated dialogue, which reflects the purpose of designing the interactive mode to reduce the logical flaws among multi-round conversations. Besides, we argue that doctor agent strategy alone is not sufficient to generate the optimal candidate. As pointed out by \cite{NEURIPS2023_271db992}, the autoregressive mechanism of LLM for generating text confines the candidate decisions by its token-level decision and left-to-right fashion. Thus, our results indicate the importance of the proposed candidate ranking strategy, which overcomes the limitation of the autoregressive mechanism by searching and ranking among a larger pool of candidates.

\subsection{Demonstration of proactive query generation}
In Figure \ref{demo}, we demonstrate four representative cases of single-round dialogue between the patient agent and doctoral agents. Note that all the examples shown in the figure are in the first round of conversation, where two variations of our method (doctor agent+candidate ranking) and (doctor agent+candidate ranking+patient agent) generate the same response. In case (a), our doctor agent+candidate ranking model actively asks for the discomfort (symptoms) of the patient, which is in accord with the real-world dialogue. In case (b), the doctor agent+ranking model further asks for more details about the irregular heartbeat (symptoms) and previous medical testing records (medical evidence data). In case (c), the doctor agent+ranking model questions the symptoms of the patient. In case (d), the doctor agent+ranking model refers to the cardio ultrasound testing (medical testing records). In contrast, the doctor agent model turns to directly answering the query of the patient agent by explaining medical terminologies. The demonstrations indicate that the candidate ranking strategy further boosts the capability of LLMs to proactively ask for and collect diagnostic information as proactive dialogue generator. 

\section{Discussion}

The experimental results confirm that the proposed doctor agent, candidate ranking strategy, and patient agent provide clinical dialogues that mimic the clinical conversation. \textcolor{black}{Moreover, our proposed model does not adopt any prior knowledge/assumption of any disease or language. As such, it is generic for different types of diseases and languages.}
Benefiting from the doctor agent and candidate ranking strategies, the proposed methods achieve higher BLEU and ROUGE scores, as well as high-level metrics such as Fluency, Professionalism, and Safety. Moreover, the interactive patient agent improves the high-level scores among multi-round conversations. Although these high-level metrics are not a part of the ranking criteria, the improved high-level scores demonstrate the effectiveness and robustness of the candidate ranking strategy. Also, we confirm that the generated dialogues by the proactive dialogue generator can provide comprehensive diagnostic information compared to the real-world dataset. 
It is worth mentioning that in real clinical practice, there should be a patient who addresses the queries from the doctor's agent. Limited by the availability of the dataset and patients' involvement, we developed the interactive patient agent to answer queries from the doctor agent and ask follow-up questions. As suggested by the experimental results, the interactive patient agent reduces the inconsistency and flaws in logic in the multi-round conversations. 
\textcolor{black}{Also, We observe a further increase in fluency, professionalism, and safety scores, after the patient agent is finetuned. }

The outcomes of our framework have multiple applications. For example, the generated dialogue can be employed to train natural language models for extracting symptoms and diagnosing diseases. The doctor agent can be used to generate query to patient and extracting diagnostic information in clinics. Additionally, it can be used for educational purposes, such as training avatars to interact with healthcare professional students in the role of standardized patients during clinical examinations. Also, the patient agent can be used for training purposes for clinical professionals to practice clinical interactions.

Beyond the proposed framework, we believe that more agents can be potentially incorporated into the framework. For example, nursing agents, if fine-tuned in different domains of medicine, can be added to our system to provide detailed suggestions which is tailored by the patients' needs. Also, a supervision agent, focusing on the correctness of the doctor agent's output, can be added to the framework to further improve the quality of the generated dialogues. With incremental data, we can also train a model for specialist, such as cardiologist, neurologist, etc., in clinical diagnosis.

\vspace{-10pt}
\section{Conclusion}
In this paper, we develop a diagnostic system to proactively collect diagnostic information via interactive conversations between doctor and patient agents. Using the proposed doctor agent, two-stage recommendation structure, and interactive patient agent, we perform comprehensive experiments on a real-world medical dialogue dataset. The BLEU and ROUGE scores show that the proposed method, after fine-tuning and candidate ranking, better mimics the conversation style in real-world dialogue. Moreover, the proposed framework achieves better performance in terms of high-level metrics including Fluency, Professionalism, and Safety. The generated dialogues are capable of providing diagnostic information.
\textcolor{black}{In the future, we will conduct a human evaluation to further evaluate our model in real-world settings. Specifically, we will invite patients and physicians to evaluate the performance of the proposed algorithms from multiple aspects, including friendliness, efficiency, and accuracy, over the conversation.}

\section{Acknowledgment}
Financial support for this publication partially results from Scialog grant \#SA-AUT-2024-022b from Research Corporation for Science Advancement and Arnold and Mabel Beckman Foundation.

\bibliography{mybibliography}{}

\begin{thebibliography}{38}
\providecommand{\natexlab}[1]{#1}
\providecommand{\url}[1]{\texttt{#1}}
\expandafter\ifx\csname urlstyle\endcsname\relax
  \providecommand{\doi}[1]{doi: #1}\else
  \providecommand{\doi}{doi: \begingroup \urlstyle{rm}\Url}\fi

\bibitem[Trevena et~al.(2006)Trevena, BPsych, Barratt, et~al.]{trevena2006systematic}
Lyndal~J Trevena, Heather M~Davey BPsych, Alexandra Barratt, et~al.
\newblock A systematic review on communicating with patients about evidence.
\newblock \emph{Journal of evaluation in clinical practice}, 12\penalty0 (1):\penalty0 13--23, 2006.

\bibitem[G{\o}tzsche(2008)]{gotzsche2008rational}
Peter G{\o}tzsche.
\newblock \emph{Rational diagnosis and treatment: evidence-based clinical decision-making}.
\newblock John Wiley \& Sons, 2008.

\bibitem[Brunetti et~al.(2019)Brunetti, Carnimeo, Trotta, et~al.]{brunetti2019computer}
Antonio Brunetti, Leonarda Carnimeo, Gianpaolo~Francesco Trotta, et~al.
\newblock Computer-assisted frameworks for classification of liver, breast and blood neoplasias via neural networks: A survey based on medical images.
\newblock \emph{Neurocomputing}, 335:\penalty0 274--298, 2019.

\bibitem[Zhao et~al.(2023)Zhao, Wang, Gu, et~al.]{zhao2023chatcad+}
Zihao Zhao, Sheng Wang, Jinchen Gu, et~al.
\newblock Chat{CAD}+: Towards a universal and reliable interactive {CAD} using {LLM}s.
\newblock \emph{arXiv preprint arXiv:2305.15964}, 2023.

\bibitem[Zhou et~al.(2023)Zhou, He, Sun, et~al.]{zhou2023skingpt4}
Juexiao Zhou, Xiaonan He, Liyuan Sun, et~al.
\newblock Skin{GPT}-4: An interactive dermatology diagnostic system with visual large language model, 2023.

\bibitem[Zhu et~al.(2023)Zhu, Chen, Shen, et~al.]{zhu2023minigpt4}
Deyao Zhu, Jun Chen, Xiaoqian Shen, et~al.
\newblock Mini{GPT}-4: Enhancing vision-language understanding with advanced large language models, 2023.

\bibitem[Chen et~al.(2021)Chen, Shen, Song, and Wan]{chen-etal-2021-cross-modal}
Zhihong Chen, Yaling Shen, Yan Song, and Xiang Wan.
\newblock Cross-modal memory networks for radiology report generation.
\newblock In \emph{Proceedings of the 59th Annual Meeting of the Association for Computational Linguistics and the 11th International Joint Conference on Natural Language Processing (Volume 1: Long Papers)}, pages 5904--5914, Online, August 2021. Association for Computational Linguistics.
\newblock \doi{10.18653/v1/2021.acl-long.459}.
\newblock URL \url{https://aclanthology.org/2021.acl-long.459}.

\bibitem[Paulhus et~al.(2007)Paulhus, Vazire, et~al.]{paulhus2007self}
Delroy~L Paulhus, Simine Vazire, et~al.
\newblock The self-report method.
\newblock \emph{Handbook of research methods in personality psychology}, 1\penalty0 (2007):\penalty0 224--239, 2007.

\bibitem[Bruce and Fries(2003)]{bruce2003stanford}
Bonnie Bruce and James~F Fries.
\newblock The stanford health assessment questionnaire: a review of its history, issues, progress, and documentation.
\newblock \emph{The Journal of rheumatology}, 30\penalty0 (1):\penalty0 167--178, 2003.

\bibitem[Stange et~al.(1998)Stange, Zyzanski, Smith, Kelly, Langa, Flocke, and Ja{\'e}n]{stange1998valid}
Kurt~C Stange, Stephen~J Zyzanski, Tracy~Fedirko Smith, Robert Kelly, Doreen~M Langa, Susan~A Flocke, and Carlos~R Ja{\'e}n.
\newblock How valid are medical records and patient questionnaires for physician profiling and health services research?: A comparison with direct observation of patient visits.
\newblock \emph{Medical care}, 36\penalty0 (6):\penalty0 851--867, 1998.

\bibitem[Bergmann et~al.(2004)Bergmann, Jacobs, Hoffmann, et~al.]{bergmann2004agreement}
Manuela~M Bergmann, Eric~J Jacobs, Kurt Hoffmann, et~al.
\newblock Agreement of self-reported medical history: comparison of an in-person interview with a self-administered questionnaire.
\newblock \emph{European journal of epidemiology}, 19:\penalty0 411--416, 2004.

\bibitem[Stirratt et~al.(2015)Stirratt, Dunbar-Jacob, Crane, Simoni, Czajkowski, Hilliard, Aikens, Hunter, Velligan, Huntley, et~al.]{stirratt2015self}
Michael~J Stirratt, Jacqueline Dunbar-Jacob, Heidi~M Crane, Jane~M Simoni, Susan Czajkowski, Marisa~E Hilliard, James~E Aikens, Christine~M Hunter, Dawn~I Velligan, Kristen Huntley, et~al.
\newblock Self-report measures of medication adherence behavior: recommendations on optimal use.
\newblock \emph{Translational behavioral medicine}, 5\penalty0 (4):\penalty0 470--482, 2015.

\bibitem[Jin et~al.(2022)Jin, Yuan, Xiong, et~al.]{jin2022biomedical}
Qiao Jin, Zheng Yuan, Guangzhi Xiong, et~al.
\newblock Biomedical question answering: a survey of approaches and challenges.
\newblock \emph{ACM Computing Surveys (CSUR)}, 55\penalty0 (2):\penalty0 1--36, 2022.

\bibitem[Lewis et~al.(2020)Lewis, Perez, Piktus, et~al.]{10.5555/3495724.3496517}
Patrick Lewis, Ethan Perez, Aleksandra Piktus, et~al.
\newblock Retrieval-augmented generation for knowledge-intensive nlp tasks.
\newblock In \emph{Proceedings of the 34th International Conference on Neural Information Processing Systems}, NIPS '20, Red Hook, NY, USA, 2020. Curran Associates Inc.
\newblock ISBN 9781713829546.

\bibitem[Pinho et~al.(2017)Pinho, Godinho, Valente, and Costa]{pinho2017multimodal}
Eduardo Pinho, Tiago Godinho, Frederico Valente, and Carlos Costa.
\newblock A multimodal search engine for medical imaging studies.
\newblock \emph{Journal of digital imaging}, 30:\penalty0 39--48, 2017.

\bibitem[Chowdhery et~al.(2024)Chowdhery, Narang, Devlin, et~al.]{10.5555/3648699.3648939}
Aakanksha Chowdhery, Sharan Narang, Jacob Devlin, et~al.
\newblock Palm: scaling language modeling with pathways.
\newblock \emph{J. Mach. Learn. Res.}, 24\penalty0 (1), mar 2024.
\newblock ISSN 1532-4435.

\bibitem[Touvron et~al.(2023{\natexlab{a}})Touvron, Lavril, Izacard, et~al.]{Touvron2023LLaMAOA}
Hugo Touvron, Thibaut Lavril, Gautier Izacard, et~al.
\newblock Llama: Open and efficient foundation language models.
\newblock \emph{ArXiv}, abs/2302.13971, 2023{\natexlab{a}}.
\newblock URL \url{https://api.semanticscholar.org/CorpusID:257219404}.

\bibitem[Touvron et~al.(2023{\natexlab{b}})Touvron, Martin, Stone, et~al.]{Touvron2023Llama2O}
Hugo Touvron, Louis Martin, Kevin~R. Stone, et~al.
\newblock Llama 2: Open foundation and fine-tuned chat models.
\newblock \emph{ArXiv}, abs/2307.09288, 2023{\natexlab{b}}.
\newblock URL \url{https://api.semanticscholar.org/CorpusID:259950998}.

\bibitem[AI@Meta(2024)]{llama3modelcard}
AI@Meta.
\newblock Llama 3 model card.
\newblock 2024.
\newblock URL \url{https://github.com/meta-llama/llama3/blob/main/MODEL_CARD.md}.

\bibitem[Brown et~al.(2020)Brown, Mann, Ryder, et~al.]{10.5555/3495724.3495883}
Tom~B. Brown, Benjamin Mann, Nick Ryder, et~al.
\newblock In \emph{Proceedings of the 34th International Conference on Neural Information Processing Systems}, NIPS '20, Red Hook, NY, USA, 2020. Curran Associates Inc.
\newblock ISBN 9781713829546.

\bibitem[OpenAI et~al.(2024)OpenAI, Achiam, Adler, et~al.]{openai2024gpt4technicalreport}
OpenAI, Josh Achiam, Steven Adler, et~al.
\newblock Gpt-4 technical report, 2024.
\newblock URL \url{https://arxiv.org/abs/2303.08774}.

\bibitem[Du et~al.(2022)Du, Qian, Liu, et~al.]{du-etal-2022-glm}
Zhengxiao Du, Yujie Qian, Xiao Liu, et~al.
\newblock {GLM}: General language model pretraining with autoregressive blank infilling.
\newblock In Smaranda Muresan, Preslav Nakov, and Aline Villavicencio, editors, \emph{Proceedings of the 60th Annual Meeting of the Association for Computational Linguistics (Volume 1: Long Papers)}, pages 320--335, Dublin, Ireland, May 2022. Association for Computational Linguistics.
\newblock \doi{10.18653/v1/2022.acl-long.26}.
\newblock URL \url{https://aclanthology.org/2022.acl-long.26}.

\bibitem[Zeng et~al.(2022)Zeng, Liu, Du, et~al.]{Zeng2022GLM130BAO}
Aohan Zeng, Xiao Liu, Zhengxiao Du, et~al.
\newblock Glm-130b: An open bilingual pre-trained model.
\newblock \emph{ArXiv}, abs/2210.02414, 2022.
\newblock URL \url{https://api.semanticscholar.org/CorpusID:252715691}.

\bibitem[Singhal et~al.(2023)Singhal, Tu, Gottweis, et~al.]{singhal2023expertlevelmedicalquestionanswering}
Karan Singhal, Tao Tu, Juraj Gottweis, et~al.
\newblock Towards expert-level medical question answering with large language models, 2023.
\newblock URL \url{https://arxiv.org/abs/2305.09617}.

\bibitem[Chen et~al.(2023)Chen, Cano, Romanou, et~al.]{chen2023meditron70bscalingmedicalpretraining}
Zeming Chen, Alejandro~Hernández Cano, Angelika Romanou, et~al.
\newblock Meditron-70b: Scaling medical pretraining for large language models, 2023.
\newblock URL \url{https://arxiv.org/abs/2311.16079}.

\bibitem[Wu et~al.(2023{\natexlab{a}})Wu, Lin, Zhang, et~al.]{wu2023pmcllamabuildingopensourcelanguage}
Chaoyi Wu, Weixiong Lin, Xiaoman Zhang, et~al.
\newblock Pmc-llama: Towards building open-source language models for medicine, 2023{\natexlab{a}}.
\newblock URL \url{https://arxiv.org/abs/2304.14454}.

\bibitem[Zhang et~al.(2024)Zhang, Yu, Adhikarla, et~al.]{zhang2024biomedgptunifiedgeneralistbiomedical}
Kai Zhang, Jun Yu, Eashan Adhikarla, et~al.
\newblock Biomedgpt: A unified and generalist biomedical generative pre-trained transformer for vision, language, and multimodal tasks, 2024.
\newblock URL \url{https://arxiv.org/abs/2305.17100}.

\bibitem[OpenAI(2023)]{openai2023gpt}
R~OpenAI.
\newblock Gpt-4 technical report. arxiv 2303.08774.
\newblock \emph{View in Article}, 2:\penalty0 13, 2023.

\bibitem[Hoffmann et~al.(2022)Hoffmann, Borgeaud, Mensch, et~al.]{hoffmann2022training}
Jordan Hoffmann, Sebastian Borgeaud, Arthur Mensch, et~al.
\newblock Training compute-optimal large language models.
\newblock \emph{arXiv preprint arXiv:2203.15556}, 2022.

\bibitem[Touvron et~al.(2023{\natexlab{c}})Touvron, Martin, Stone, et~al.]{touvron2023llama}
Hugo Touvron, Louis Martin, Kevin Stone, et~al.
\newblock Llama 2: Open foundation and fine-tuned chat models, 2023{\natexlab{c}}.

\bibitem[Chowdhery et~al.(2023)Chowdhery, Narang, Devlin, et~al.]{chowdhery2023palm}
Aakanksha Chowdhery, Sharan Narang, Jacob Devlin, et~al.
\newblock Palm: Scaling language modeling with pathways.
\newblock \emph{Journal of Machine Learning Research}, 24\penalty0 (240):\penalty0 1--113, 2023.

\bibitem[Wu et~al.(2023{\natexlab{b}})Wu, He, Liu, et~al.]{10113601}
Tianyu Wu, Shizhu He, Jingping Liu, et~al.
\newblock A brief overview of chatgpt: The history, status quo and potential future development.
\newblock \emph{IEEE/CAA Journal of Automatica Sinica}, 10\penalty0 (5):\penalty0 1122--1136, 2023{\natexlab{b}}.
\newblock \doi{10.1109/JAS.2023.123618}.

\bibitem[Zhang et~al.(2023)Zhang, Chen, Jiang, et~al.]{zhang-etal-2023-huatuogpt}
Hongbo Zhang, Junying Chen, Feng Jiang, et~al.
\newblock {H}uatuo{GPT}, towards taming language model to be a doctor.
\newblock In Houda Bouamor, Juan Pino, and Kalika Bali, editors, \emph{Findings of the Association for Computational Linguistics: EMNLP 2023}, pages 10859--10885, Singapore, December 2023. Association for Computational Linguistics.
\newblock \doi{10.18653/v1/2023.findings-emnlp.725}.
\newblock URL \url{https://aclanthology.org/2023.findings-emnlp.725}.

\bibitem[Yang et~al.(2023)Yang, Zhao, Zhu, et~al.]{yang2023zhongjing}
Songhua Yang, Hanjia Zhao, Senbin Zhu, et~al.
\newblock Zhongjing: Enhancing chinese medical capabilities of large language models through expert feedback and real-world multi-turn dialogues.
\newblock \emph{arXiv preprint arXiv:2308.03549}, 2023.

\bibitem[Zhang et~al.(2020)Zhang, Jiang, Zhang, et~al.]{Zhang2020MIEAM}
Yuanzhe Zhang, Zhongtao Jiang, Tao Zhang, et~al.
\newblock Mie: A medical information extractor towards medical dialogues.
\newblock In \emph{Annual Meeting of the Association for Computational Linguistics}, 2020.
\newblock URL \url{https://api.semanticscholar.org/CorpusID:220047186}.

\bibitem[Papineni et~al.(2002)Papineni, Roukos, Ward, and Zhu]{papineni-etal-2002-bleu}
Kishore Papineni, Salim Roukos, Todd Ward, and Wei-Jing Zhu.
\newblock {B}leu: a method for automatic evaluation of machine translation.
\newblock In Pierre Isabelle, Eugene Charniak, and Dekang Lin, editors, \emph{Proceedings of the 40th Annual Meeting of the Association for Computational Linguistics}, pages 311--318, Philadelphia, Pennsylvania, USA, July 2002.
\newblock \doi{10.3115/1073083.1073135}.
\newblock URL \url{https://aclanthology.org/P02-1040}.

\bibitem[Lin(2004)]{lin-2004-rouge}
Chin-Yew Lin.
\newblock {ROUGE}: A package for automatic evaluation of summaries.
\newblock In \emph{Text Summarization Branches Out}, pages 74--81, Barcelona, Spain, July 2004. Association for Computational Linguistics.
\newblock URL \url{https://aclanthology.org/W04-1013}.

\bibitem[Yao et~al.(2023)Yao, Yu, Zhao, et~al.]{NEURIPS2023_271db992}
Shunyu Yao, Dian Yu, Jeffrey Zhao, et~al.
\newblock Tree of thoughts: Deliberate problem solving with large language models.
\newblock In A.~Oh, T.~Naumann, A.~Globerson, K.~Saenko, M.~Hardt, and S.~Levine, editors, \emph{Advances in Neural Information Processing Systems}, volume~36, pages 11809--11822. Curran Associates, Inc., 2023.
\newblock URL \url{https://proceedings.neurips.cc/paper_files/paper/2023/file/271db9922b8d1f4dd7aaef84ed5ac703-Paper-Conference.pdf}.

\end{thebibliography}

\end{document}